\def\year{2019}\relax
\documentclass[letterpaper]{article} 
\usepackage{aaai19}  
\usepackage{times}  
\usepackage{helvet}  
\usepackage{courier}  
\usepackage{url}  
\usepackage{graphicx}  
\usepackage{amssymb}
\usepackage{graphbox}
\usepackage{enumitem}
\usepackage{subcaption} 
\begin{document}
%
\title{Depthwise Convolution is All You Need for Learning Multiple Visual Domains}
\author{Yunhui Guo $^{\dagger}$ $^{*}$ \quad  Yandong Li $^{\ddagger}$ $^{*}$ 
\quad  Rogerio Feris $^{\dagger\dagger}$  \quad Liqiang Wang$^{\ddagger}$ \quad Tajana Rosing $^{\dagger}$ \\
{University of California, San Diego, CA} $^{\dagger}$  
\quad IBM Research AI $^{\dagger\dagger}$
\quad University of Central Florida,
Orlando, FL $^{\ddagger}$  \\
yug185@eng.ucsd.edu, lyndon.leeseu@outlook.com,
rsferis@us.ibm.com,
lwang@cs.ucf.edu,
tajana@ucsd.edu}
\maketitle
\begin{abstract}
There is a growing interest in designing models that can deal with images from different visual domains. If there exists a universal structure in different visual domains that can be captured via a common parameterization, then we can use a single model for all domains rather than one model per domain. A model aware of the relationships between different
domains can also be trained to work on new domains
with less resources. However, to identify the reusable structure in a model is not easy. In this paper, we propose a multi-domain learning architecture based on depthwise separable convolution. The proposed approach is based on the assumption that images from different domains share cross-channel correlations but have domain-specific spatial correlations. The proposed model is compact and has minimal overhead when being applied to new domains. Additionally, we introduce a gating mechanism to promote soft sharing between different domains. We evaluate our approach on Visual Decathlon Challenge, a benchmark for testing the ability of multi-domain models. The experiments show that our approach can achieve the highest score while only requiring 50\% of the parameters compared with the state-of-the-art approaches.
\end{abstract}
\let\thefootnote\relax\footnote{$^{*}$ Equal contribution. Work done during internship at IBM Research mentored by Rogerio Feris. $^{\ddagger}$ The authors' work was supported in part by NSF-1741431.}

\section{Introduction}
Deep convolutional neural networks (CNN) \cite{krizhevsky2012imagenet,he2016deep} have been the state-of-the-art methods for tackling vision tasks. The existing CNN models are powerful but mostly designed for dealing with images from a specific visual domain ({\it e.g.} digits, animals, or flowers)~\cite{Li_2018_ECCV,Gan_2017_ICCV,long2018multimodal}. This limits the applications of current approaches, as each time the network needs to be retrained when new tasks arrive. In sharp contrast to such CNN models, humans can easily generalize to new domains based on the acquired knowledge \cite{cichon2015branch,hayashi2015labelling,kirkpatrick2017overcoming,li2017learning}. Previous works \cite{bilen2017universal,rebuffi18efficient} show that images from different domains may have a universal structure that can be captured via a common parameterization. A natural question then arises:
\begin{center}
    \textit{Can we build a single neural network that can deal with images across different domains?}
\end{center}

The question motivates the field called multi-domain learning, where we target designing a common feature extractor that can capture the universal structure in different domains and reducing the overhead of adding new tasks to the model. With multi-domain learning, the visual models are vested with the ability to work well on different domains with minimal or no domain-specific parameters. 

\begin{figure}
\centering
\begin{subfigure}{0.1\textwidth}
   \includegraphics[width=1.6cm, height=1.6cm]{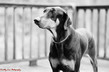}
   \caption{Animals} \label{fig:x_a}
\end{subfigure}
\begin{subfigure}{0.1\textwidth}
   \includegraphics[width=1.6cm, height=1.6cm]{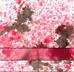}
   \caption{Textures} \label{fig:x_a}
\end{subfigure}
\begin{subfigure}{0.1\textwidth}
   \includegraphics[width=1.6cm, height=1.6cm]{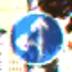}
   \caption{Signs} \label{fig:x_a}
\end{subfigure}
\begin{subfigure}{0.1\textwidth}
   \includegraphics[width=1.6cm, height=1.6cm]{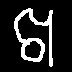}
   \caption{Omniglot} \label{fig:x_a}
\end{subfigure}

\bigskip
\begin{subfigure}{0.1\textwidth}
   \includegraphics[width=1.6cm, height=1.6cm]{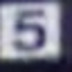}
   \caption{Digits} \label{fig:x_a}
\end{subfigure}
\begin{subfigure}{0.1\textwidth}
   \includegraphics[width=1.6cm, height=1.6cm]{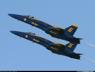}
   \caption{Aircraft} \label{fig:x_a}
\end{subfigure}
\begin{subfigure}{0.1\textwidth}
   \includegraphics[width=1.6cm, height=1.6cm]{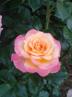}
   \caption{Flowers} \label{fig:x_a}
\end{subfigure}
\begin{subfigure}{0.1\textwidth}
   \includegraphics[width=1.6cm, height=1.6cm]{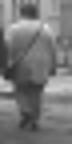}
   \caption{Pedestrian} \label{fig:x_a}
\end{subfigure}

\caption{Image examples from different domains. }
\label{figs: domains}
\end{figure}

There are two challenges in multi-domain learning. The first one is to identify a common structure among different domains. As shown in Fig \ref{figs: domains}, images from different domains are visually different, it is challenging to design a single feature extractor for all domains. Another challenge is to add new tasks to the model without introducing additional parameters. Existing neural network based multi-domain learning approaches \cite{bilen2017universal,rebuffi2017learning,rebuffi18efficient,rosenfeld2017incremental} mostly focus on the architecture design while ignoring the structural regularity hidden in different domains which leads to sub-optimal solutions.

In this paper, we propose a multi-domain learning approach based on depthwise separable convolution. Depthwise separable convolution has been proved to be a powerful variation of standard convolution for many applications, such as image classification \cite{chollet2017xception}, natural language processing \cite{kaiser2017depthwise} and embedded vision applications \cite{howard2017mobilenets}. To the best of our knowledge, this is the first work that explores depthwise separable convolution for multi-domain learning. The proposed multi-domain learning model is compact and easily extensible. To promote knowledge transfer between different domains we further introduce a softmax gating mechanism. We evaluate our method on Visual Decathlon Challenge \cite{rebuffi2017learning}, a benchmark for testing multi-domain learning models. Our method can beat the state-of-the-art models with only 50\% of the parameters.
\\

\noindent\textbf{Summary and contributions}: The contributions of this paper are summarized below:
\begin{itemize}
    \item We propose a novel multi-domain learning approach by exploiting the structure regularity hidden in different domains. The proposed approach greatly reduces the number of parameters and can be easily adapted to work on new domains.
    
    \item The proposed approach is based on the assumption that images in different domains share cross-channel correlations while having domain-specific spatial correlations. We validate the assumption by analyzing the visual concepts captured by depthwise separable convolution using \textit{network dissection} \cite{bau2017network}.

    \item Our approach outperforms the state-of-the-art results on Visual Decathlon Challenge with only 50\% of the parameters.
    
\end{itemize}

\section{Related Work}

\noindent\textbf{Multi-Domain Learning} Multi-domain learning aims at creating a single neural network to perform image classification tasks in a variety of domains. \cite{bilen2017universal} showed that a single neural network can learn simultaneously several different visual domains by using an instance normalization layer. \cite{rebuffi2017learning,rebuffi18efficient} proposed universal parametric families of neural networks that contain specialized problem-specific models which differ only by a small number of parameters. \cite{rosenfeld2017incremental} proposed a method called Deep Adaptation Networks (DAN) that constrains newly learned filters for new domains to be linear combinations of existing ones. Multi-domain learning can promote the application of deep learning based vision models since it reduces engineers' effort to train new models for new images.
\\

\noindent\textbf{Multi-Task Learning} The goal of multi-task learning \cite{bilen2016integrated,doersch2017multi,kokkinos2017ubernet,wang2017transitive} is to extract different features from a single input to simultaneously perform classification, object recognition, edge detection, etc. Various applications can be benefited from a multi-task learning approach since the training signals can be reused among related tasks \cite{caruana1997multitask,zamir2018taskonomy}. 
\\

\noindent\textbf{Transfer Learning} The goal of transfer learning is to improve the performance of a model on a target domain by leveraging the information from a related source domain \cite{pan2010survey,bengio2012deep,hu2015deep}. Transfer learning has wide applications in a variety of areas, such as computer vision \cite{raina2007self}, sentiment analysis \cite{glorot2011domain} and recommender systems 
\cite{pan2010transfer,guo2015crorank}. Different from transfer learning, multi-domain learning aims at maximizing the performance of the model across multiple domains rather than focusing on a specific target domain.
\\


\section{Preliminary}
 \begin{figure*}[t]
\centering
\includegraphics[width=1.0\linewidth]{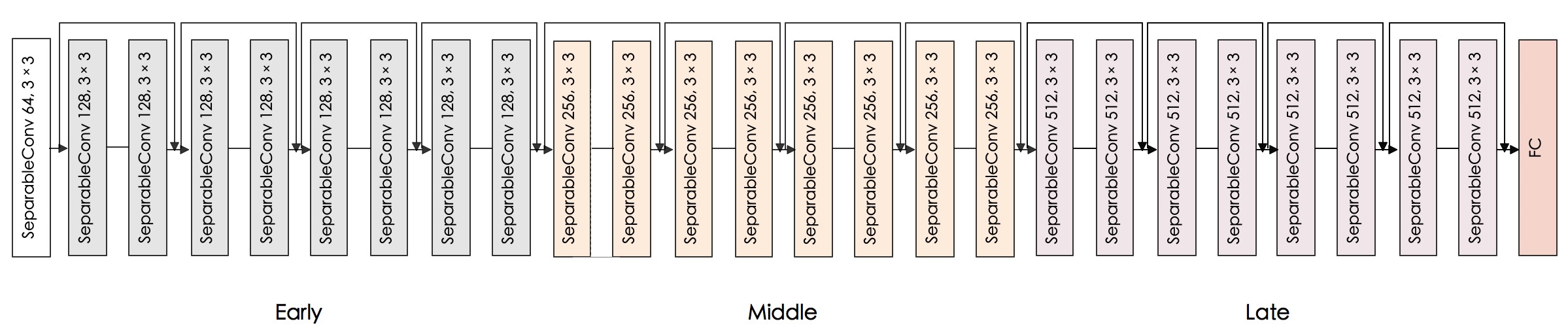}
\caption{ResNet-26 with depthwise separable convolution.}
\label{fig: arch}
\end{figure*}

\begin{figure}[h]
\centering
     \includegraphics[width=0.13\textwidth]{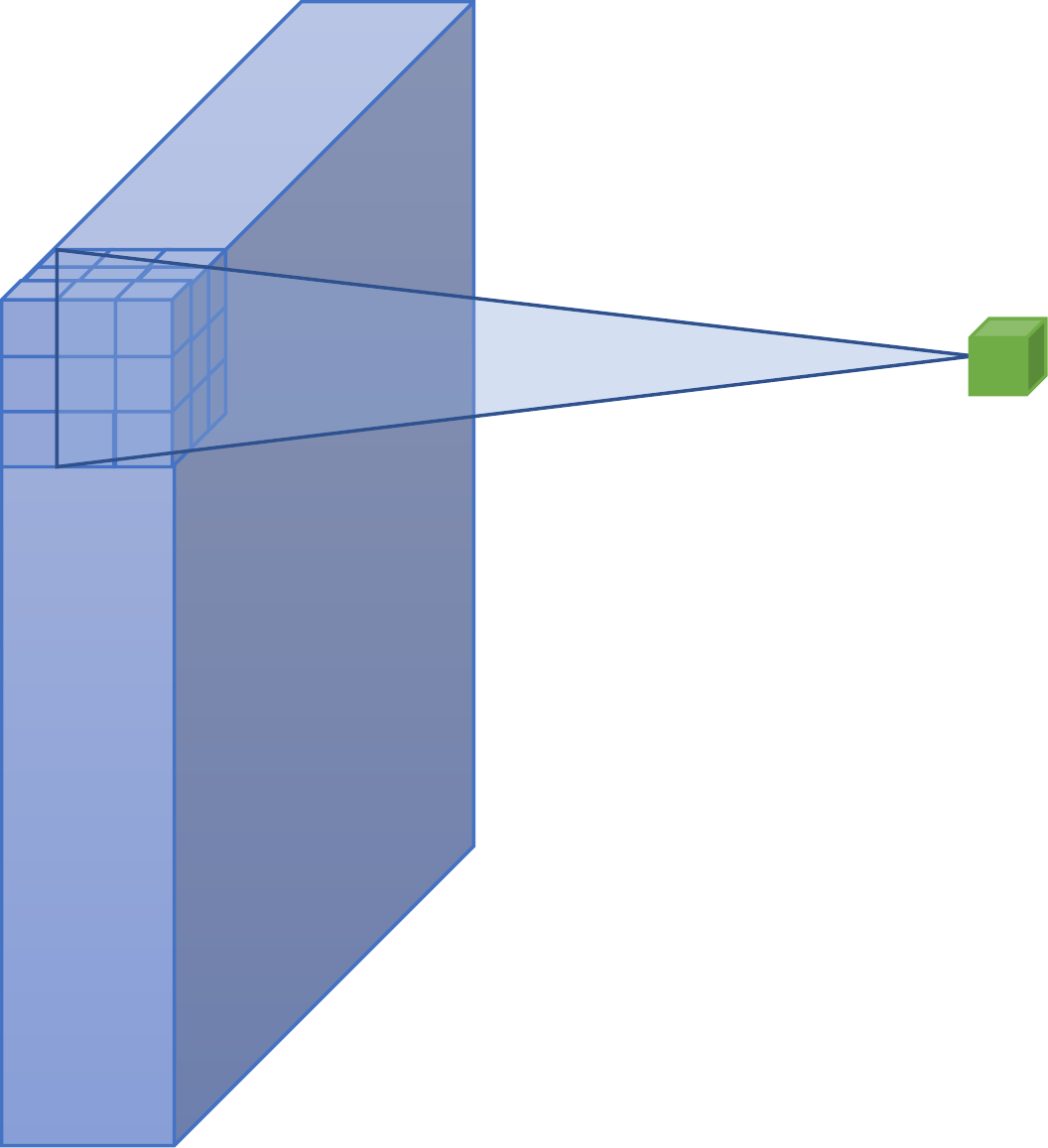}
     \hfil
     \includegraphics[width=0.2\textwidth]{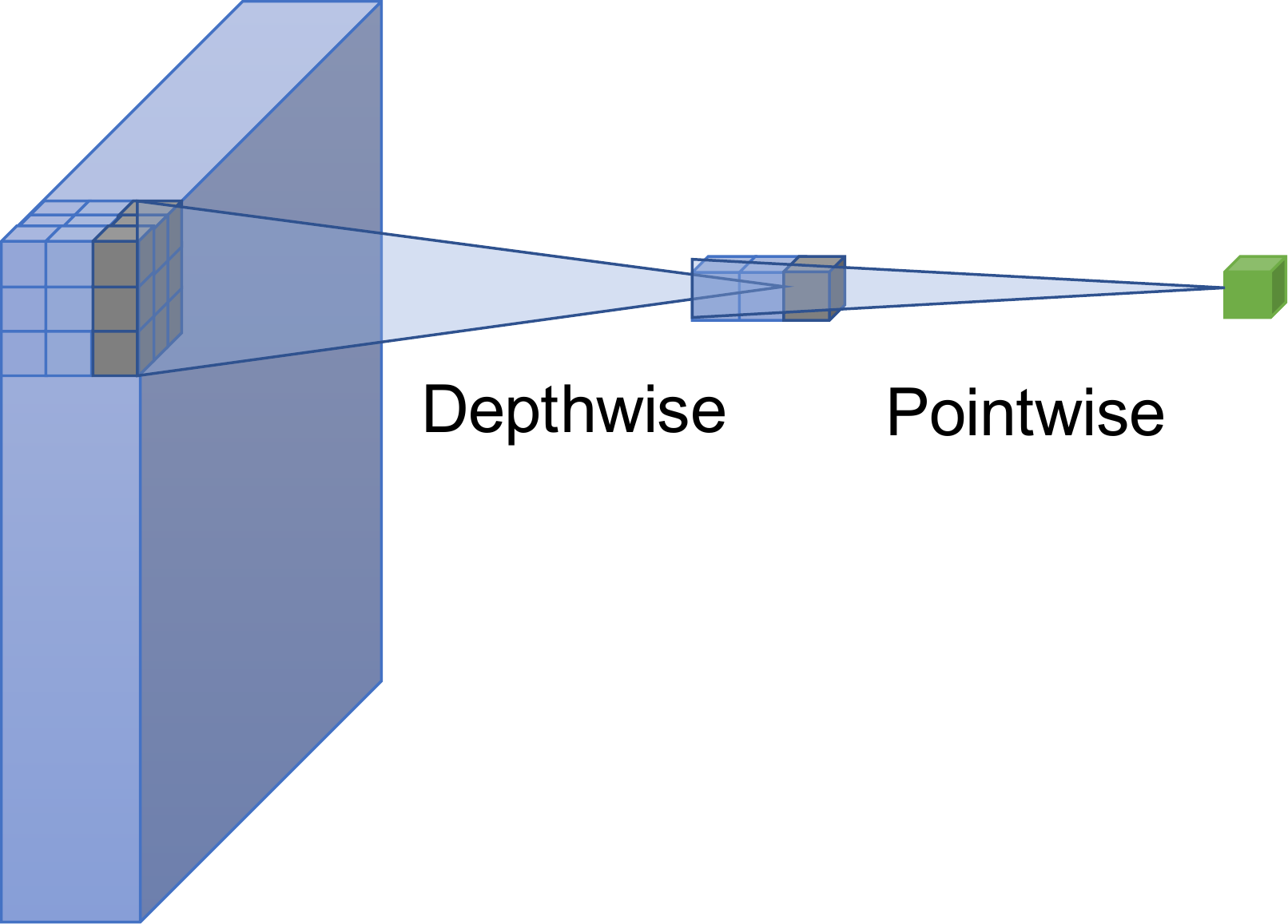}
\caption{Standard convolution and depthwise separable convolution.}   \label{fig: convs}
\end{figure}

\subsubsection{Problem Definition and Notations} ~\\
\noindent Consider a set of image domains $\{D_1, D_2, ..., D_T\}$, each domain $D_i$ consists of a triplet $\{X_i, Y_i, P_i\}$. $X_i \in \mathbb{R}^{C_i \times H_i\times W_i}$ is the input image space and $Y_i \in \{1,2,...,L_i\}$ is the output label space. Let $x \in X_i$ and $y \in Y_i$ be a pair of objects. The joint probabilistic distribution $P_i(x,y)$ describes the frequency of encountering $(x, y)$ in domain $D_i$. For a neural network $g_i(x)$: $\mathbb{R}^{C_i \times H_i\times W_i} \rightarrow \{1,2,...,L_i\}$ and a given loss function $l$, the risk of $g_i(x)$ can be measured as below,

\begin{equation}
    R_i = \mathbb{E}[l(y,g_i(x)] = \int l(y,g_i(x))d P_i(x,y)
\end{equation}

In multi-domain learning, our goal is to design neural network architectures that can work well on all the domains simultaneously. Let $\mathcal{E}_{(D_i)}$ be the domain-specific parameters for domain $D_i$ and $\mathcal{C}$ be the sharable portion of the neural network. For $x \in X_i$, the output of the network can be calculated as,
\begin{equation}
    \hat{y} = (\mathcal{E}_{(D_i)} \circ \mathcal{C})(x)
\end{equation}

The average risk of the neural network across all the domains can be expressed as,
\begin{equation}
    R = \frac{1}{T}\sum_{i=1}^{T} \mathbb{E}[l(y,(\mathcal{E}_{(D_i)} \circ \mathcal{C})(x)]
\end{equation}
The goals of multi-domain learning include: (1) minimize the average risk across different domains; (2) maximize the size of sharing part $\mathcal{C}$; (3) minimize the size of the domain-specific part $\mathcal{E}_{(D_i)}$.

\subsection{Depthwise Separable Convolution}
Our proposed approach is based on depthwise separable convolution that factorizes a standard $3 \times 3$ convolution into a $3 \times 3$ depthwise
convolution and a $1\times1$ pointwise convolution. While standard convolution performs the channel-wise and spatial-wise computation in one step, depthwise separable convolution splits the computation into two steps: depthwise convolution applies a single convolutional filter per each input channel and pointwise convolution is used to create a linear combination of the output of the depthwise convolution. The comparison of standard convolution and depthwise separable convolution is shown in Fig. \ref{fig: convs}.

Consider applying a standard convolutional filter $K$ of size $W \times W \times M \times N$ on an input feature map $F$ of size $D_f \times D_f \times M$ and produces an output feature map $O$ is of size $D_f \times D_f \times N$,

\begin{equation}
    O_{k,l,n} = \sum_{i,j,m}K_{i,j,m,n} \cdot F_{k+i-1, l+j-1,m}
\end{equation}

In depthwise separable convolution, we factorize above computation into two steps. The first step applies a $3 \times 3$ depthwise
convolution $\hat{K}$ to each input channel,
\begin{equation}
    \hat{O}_{k,l,m} = \sum_{i,j}\hat{K}_{i,j,m} \cdot F_{k+i-1, l+j-1,m}
\end{equation}

The second step applies $1 \times 1$ pointwise
convolution $\tilde{K}$ to combine the output of depthwise convolution,

\begin{equation}
    O_{k,l,n} = \sum_{m}\tilde{K}_{m,n} \cdot \hat{O}_{k-1, l-1,m}
\end{equation}
\\

Depthwise convolution and pointwise convolution have different roles in generating new features: the former is used for capturing spatial correlations while the latter is used for capturing channel-wise correlations.

Most the previous works \cite{chollet2017xception,howard2017mobilenets,sandler2018inverted} focus on the computational aspect of depthwise separable convolution since it requires less parameters than standard convolution and is more computationally effective. In \cite{chollet2017xception}, the authors proposed the ``Inception hypothesis" stating that mapping cross-channel correlations and spatial correlations separately is more efficient than mapping them at once. In this paper, we provide further evidence to support this hypothesis in the setting of multi-domain learning. We validate the assumption that images from different domains share cross-channel correlations but have domain-specific spatial correlations. Based on this idea, we develop a highly efficient multi-domain learning method. We further analyze the visual concepts captured by depthwise convolution and pointwise convolution based on \textit{network dissection} \cite{bau2017network}. The visualization results show that while having less parameters depthwise convolution captures more concepts than pointwise convolution.

\section{Proposed Approach}
\subsection{Network Architecture}
For the experiments, we use the same ResNet-26 architecture as in \cite{rebuffi18efficient}. This allows us to fairly compare the performance of the proposed approach with previous ones. This original architecture has three macro residual blocks, each outputting 64, 128, 256 feature channels. Each macro block consists of 4 residual blocks. Each residual block has two convolutional layers consisting of 3 $\times$ 3 convolutional filters. The network ends with a global average pooling layer and a softmax layer for classification. 

Different from \cite{rebuffi18efficient}, we replace each standard convolution in the ResNet-26 with depthwise separable convolution and increase the channel size. The modified network architecture is shown in Fig. \ref{fig: arch}. This choice leads to a more compact model while still maintaining enough network capacity. The original ResNet-26 has over 6M parameters while our modified architecture has only half the amount of parameters. In the experiments we found that the reduction of parameters does no harm to the performance of the model. The use of depthwise separable convolution allows us to model cross-channel correlations and spatial correlations separately. The idea behind our multi-domain learning method is to leverage the different roles of cross-channel correlations and spatial correlations in generating image features by sharing the pointwise convolution across different domains. 

\begin{figure*}[t]
\centering
\includegraphics[width=0.7\linewidth]{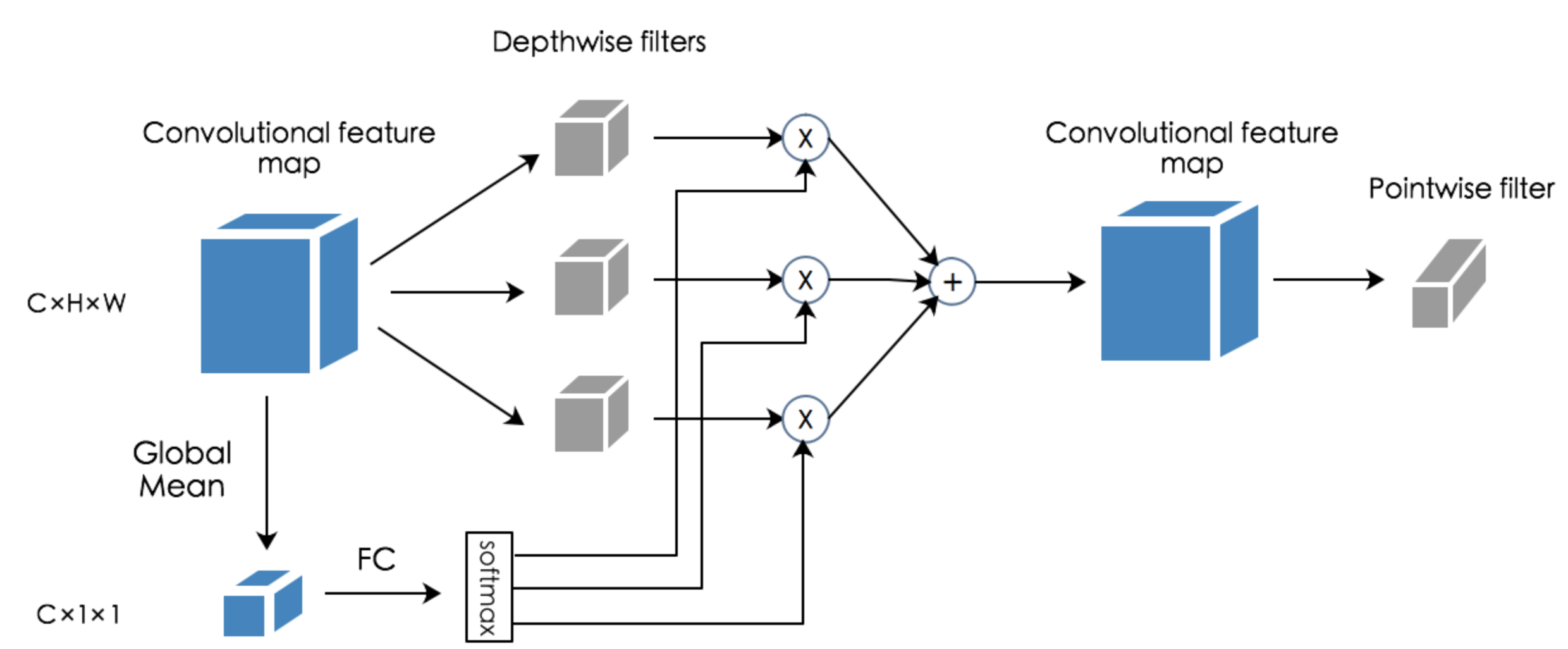}
\caption{The proposed soft-sharing approach for sharing spatial correlations.}
\label{fig: transfer}
\end{figure*}
\subsection{Learning Multiple Domains}
For multi-domain learning, it is essential to have a set of universally sharable parameters that can generalize to unseen domains. To get a good starting set of parameters, we first train the modified ResNet-26 on ImageNet. After we obtain a well-initialized network, each time when a new domain arrives, we add a new output layer and finetune the depth-wise convolutional filters. The pointwise convolutional filters are shared accross different domains. Since the statistics of the images from different domains are different, we also allow domain-specific batch normalization parameters. During inference, we stack the trained depthwise convolutional filters for all domains as a 4D tensor and the output of domain $d$ can be calculated as,
\begin{equation}
    \hat{O}_{k,l,m, d} = \sum_{i,j}\hat{K}_{i,j,m, d} \cdot F_{k+i-1, l+j-1,m,d}
\end{equation}

The adoption of depthwise separable convolution provides a natural separation for modeling cross-channel correlations and spatial correlations. Experimental evidence \cite{chollet2017xception} suggests the decouple of cross-channel correlations and spatial correlations would result in more useful features. We take one step further to develop a multi-domain domain method based on the assumption that different domains share cross-channel correlations but have domain-specific spatial correlations. Our method is based on two observations: model efficiency and interpretability of hidden units in a deep neural network.  \\

\noindent\textbf{Model efficiency} Table \ref{table: comparison} shows the comparison of standard $3 \times 3$ convolution, $3 \times 3$ depthwise convolution (Dwise) and $1 \times 1$ pointwise convolution (Pwise). Clearly, standard convolution has far more parameters than both depthwise convolution ($\times c_2$) and pointwise convolution ($\times 9$). Typically, pointwise convolution has more parameters than depthwise convolution. In the architecture shown in Fig \ref{fig: arch}, pointwise convolution accounts for 80\% of the parameters in the convolutional layers. The choice of sharing pointwise convolution and adding depthwise convolution induces minimal additional parameters when dealing with new domains. In the experiments we found that only by adding depthwise convolution leads to a network with limited number of free parameters which cannot handle some large datasets. To increase the network capacity, we allow the last convolutional layer to be specific for each domain. Based on this modification, each new domain averagely introduces 0.3M additional parameters which is 10\% of the modified ResNet-26.
\\

\noindent\textbf{Interpretability} While depthwise convolution typical has less paramaters, by using the technique of \textit{network dissection} \cite{bau2017network}, we found it captures more visual concepts than pointwise convolution. Meanwhile, the results in the same convolutional layer show that depthwise convolution captures higher level concepts such as wheel and grass while pointwise convolution can only detect dots or honeycombed. This observation suggests that pointwise convolution can be generally shared between different image domains since it is typically used for dealing with lower level features.  

\begin{center}
\begin{table}
\small
\begin{tabular}{ |c|c|c|c| } 
 \hline
Input & Operator & Output  & Parameters \\ 
\hline
 $c_1 \times h \times w$ & $3 \times 3 \textnormal{ Conv2d}$ &  $c_2 \times h \times w$ &  $3 \times 3 \times c_1 \times c_2$ \\ 
 $c_1 \times h \times w$ & $3 \times 3 \textnormal{ Dwise }$ &  $c_1 \times h \times w$ & $3 \times 3 \times c_1$ \\
  $c_1 \times h \times w$ & $1 \times 1 \textnormal{ Pwise }$ & $c_2 \times h \times w$ & $1 \times 1 \times c_1 \times c_2$ \\ 
 \hline
\end{tabular}
\caption{Comparison of standard $3 \times 3$ convolution, $3 \times 3$ depthwise convolution (Dwise) and $1 \times 1$ pointwise convolution (Pwise).}
\label{table: comparison}
\end{table}
\end{center}

\subsection{Soft Sharing of Trained Depthwise Filters}
In addition to the proposed sharing pointwise filters (cross-channel correlations) for multi-domain learning, we also investigate whether the depthwise filters (spatial correlations) learned from other domains can be transferred to the target domain. We introduce a novel soft sharing approach in the multi-domain setting to allow the sharing of depthwise convolution. We first train domain-specific depthwise filters. Then we stack all the domain-specific filters as in Fig~\ref{fig: transfer}. During soft-sharing, we train each domain one by one. All the domain-specific depthwise filters and pointwise filters (trained on ImageNet) are fixed during soft sharing. We only train the feedforward network that controls the softmax gate. For a specific target domain, the softmax gate allows a soft sharing of trained depthwise filters with other domains. It can be denoted as follows: for each domain $D_j$, consider a network with $L$ depthwise separable convolutional layers $S_1 , . . . , S_L$, the input to the pointwise convolution in layer $l$  is defined as,

\begin{equation}
    \hat{O}^l = \sum_{i=1}^{T} s^l_i \hat{O}^l_i, \quad\textnormal{with} \sum_{i=1}^T s^l_i = 1
\end{equation}

\noindent where $\hat{O}^l_i$ is the output of the depthwise convolution of domain $i$ in the layer $l$ if we use images in domain $D_j$ as input. $s^l_i$ denotes a learned scale for the depthwise convolution of domain $i$ in the layer $l$.  The scales $s_1, ..., s_T$ are the output of a softmax gate. The input to the softmax gate is the convolutional feature map $X_{l-1} \in \mathbb{R}^{C \times H \times W}$ produced by the previous layer. Similar to \cite{veit2017convolutional},  we only consider global channel-wise features. In particular, we perform global average pooling to compute channel-wise means,
\begin{equation}
    M_c = \frac{1}{H \times W} \sum_{i=1}^{H}\sum_{j=1}^{W}X_{c,i,j}
\end{equation}

The output is a 3-dimensional tensor of size $C \times 1 \times 1$. To achieve a lightweight design, we adopt a simple feedforward network consisting of two linear layers with ReLU activations to apply a nonlinear transformation on the channel-wise means and feed the output to the softmax gate. All the convolutional filters are freezed during soft sharing. The scales $s_1, ..., s_T$ and the parameters of the feedforward networks are learnt jointly via backpropagation.

It is widely believed that early layers in a convolutional neural network are used for detecting lower level features such as textures while later layers are used for detecting parts or objects. Based on this observation, we partition the network into three regions (early, middle, late) as shown in Figure \ref{fig: arch} and consider different placement of the softmax gate which allows us to compare a variety of sharing strategies.

\section{Experiment}

\subsubsection{Datasets and evaluation metrics}
We evaluate our approach on Visual Domain Decathlon Challenge \cite{rebuffi2017learning}. It is a challenge to test the ability of visual recognition algorithms to cope with images from different visual domains. There are a total of 10 datasets: (1) \textbf{ImageNet} (2) \textbf{CIFAR-100} (3) \textbf{Aircraft} (4) \textbf{Daimler pedestrian classification} (5) \textbf{Describable textures} (6) \textbf{German traffic signs} (7) \textbf{Omniglot} (8) \textbf{SVHN} (9) \textbf{UCF101 Dynamic Images} (10) 
\textbf{VGG-Flowers}. The detailed statistics of the datasets can be found at \url{http://www.robots.ox.ac.uk/~vgg/decathlon/}.

The performance is measured in terms of a single scalar score $S=\sum_{i=1}^{10}\alpha_i \textnormal{max}\{0, E_i^{\textnormal{max}}-E_i\}^{\gamma_i}$,where $   E_i = \frac{1}{D_i^{\textnormal{test}}}\sum_{(x,y) \in D_i^{\textnormal{test}}} 1\{y \neq  (\mathcal{E}_{(D_i)} \circ \mathcal{C})(x)\}$.
$E_i$ is the average test error of domain $D_i$. $E_i^{\textnormal{max}}$ is the error of a reasonable baseline algorithm. The exponent $\gamma_i$ is set to be 2 for all domains. The coefficient $\alpha_{i}$ is  $1000(E_i^{\textnormal{max}})^{-\gamma_i}$ then a perfect classifier receives 1000. The maximum score achieved across 10 domains is 10000.

\subsubsection{Baselines}
We consider the following baselines in the experiments,
\begin{enumerate}[label=\textbf{(\alph*)}]
\item \textbf{Individual Network}: The simplest baseline we consider is Individual Network. We finetune the pretrained modified ResNet-26 on each domain which leads to 10 models altogether. This approach results in the largest model size since there is no sharing between different domains.

\item \textbf{Classifier Only}: We freeze the feature extractor part of the pretrained modified ResNet-26 on ImageNet and train domain-specific classifier layer for each domain.

\item \textbf{Depthwise Sharing}: Rather than sharing pointwise convolution, we consider an alternative approach of multi-domain extension of depthwise separable convolution which shares the depthwise convolution between different domains. 

\item \textbf{Residual Adapters}: Residual Adapters \cite{rebuffi2017learning,rebuffi18efficient} are the state-of-the-art approaches for multi-domain learning which include Serial Residual Adapter \cite{rebuffi2017learning} and Parallel Residual Adapter \cite{rebuffi18efficient}. 

\item \textbf{Deep Adaptation Networks (DAN)}: In \cite{rosenfeld2017incremental} the authors propose Deep Adaptation Networks (DAN) that constrains newly learned filters for new domains to be linear combinations of existing ones via \textit{controller modules}.

\item \textbf{PiggyBack}: In \cite{mallya2018piggyback} the authors present PiggyBack for adding multiple tasks to a single network by learning domain-specific binary masks. The main idea is derived from network quantization \cite{courbariaux2016binarized,guo2018survey} and pruning. 

\end{enumerate}

\def\arraystretch{1.2}
\begin{table*}[!htb]
	\small
	\begin{center}
		\begin{tabular}{c c c c c c c c c c c c c c c} 
			
		    Model	& \#par  & ImNet  & Airc. & C100  & DPed & DTD & GTSR & Flwr & OGlt & SVHN  &UCF  & mean & S \\
			
			\# images & & 1.3m & 7k &50k& 30k& 4k &40k& 2k &26k& 70k& 9k \\
			\hline
    
    		Serial Res. Adapt. &  $2 \times$  & 59.67 & 61.87 & 81.20  & 93.88&  57.13 & 97.57 & 81.67& 89.62 & 96.13 & 50.12  & 76.89 & 2621\\
    
    		Parallel Res. Adapt. & $2 \times$  &60.32& 64.21 &81.91 &94.73 &58.83 &99.38 &84.68& 89.21& 96.54& 50.94 & 78.07 & 3412\\
    		\hline
        	DAN & $2.17 \times$ & 57.74 &  64.12 & 80.07 & 91.30 & 56.64 & 98.46 & 86.05 & 89.67 & 96.77 & 49.38 & 77.01 & 2851  \\
            \hline
    
            Piggyback & $1.28 \times$ & 57.69 &65.29 & 79.87 & 96.99& 57.45 & 97.27 & 79.09 &87.63 & 97.24 & 47.48 & 76.60 & 2838  \\
            \hline		
    		Individual Network & $5 \times$ &  63.99&  65.71&  78.26& 88.29 & 52.19 & 98.76& 83.17 &90.04 & 96.84 & 48.35& 76.56 & 2756 \\

    		Classifier Only & $ 0.6 \times$ & 63.99& 51.04 & 75.32& 94.49 & 54.21& 98.48& 84.47 & 86.66 & 95.14 & 43.75 &74.76 & 2446  \\
    		
    		Depthwise Sharing & $4 \times$ & 63.99 & 67.42& 74.46 & 95.60& 54.85 & 98.52& 87.34 & 89.88 & 96.62 &50.39 & 77.91 &  3234 \\

    		Proposed Approach & $1\times$ & 63.99 & 61.06 & 81.20 & 97.00 & 55.48 & 99.27 & 85.67 & 89.12 & 96.16 & 49.33 & 77.82 & 3507 \\
			\hline 
		\end{tabular}
	\end{center}
		\caption{ {\textnormal{Top-1 classification accuracy and the Visual Decathlon Challenge score (S) of the proposed approach and baselines. \#par is the number of parameters w.r.t. the proposed approach.
}}}
	\label{table:results}
\end{table*}

\bgroup
\def\arraystretch{1.2}
\begin{table*}[!htb]
	\small
	\begin{center}
		\begin{tabular}{c c c c c c c c c c c c c c} 
			
		    Model	 & ImNet  & Airc. & C100  & DPed & DTD & GTSR & Flwr & OGlt & SVHN  &UCF  & mean & S \\
			
			\# images & 1.3m & 7k &50k& 30k& 4k &40k& 2k &26k& 70k& 9k \\
			\hline
    
    		early  &  63.99& 58.69  & 81.01  & 95.44 & 55.75 &  98.75& 84.90 & 88.80  & 96.18  & 48.86  & 77.23 & 3102\\
    
    		 middle  &  63.99& 59.11  & 80.93  & 95.33 & 54.74 &  98.71&  85.42& 88.93  & 96.09  &  48.91 & 77.21 & 3086\\
    		
    		 late  &  63.99&  58.81  & 80.93 & 96.63 & 54.74  & 98.91  & 84.79  & 89.35 &  96.30 & 49.01 & 77.88 & 3303 \\
			\hline 
		\end{tabular}
	\end{center}
		\caption{ {\textnormal{Top-1 classification accuracy and the Visual Decathlon Challenge score (S) of different soft sharing strategies.}}}
	\label{Table: transfer}
\end{table*}
\subsubsection{Implementation details}
All networks were implemented using Pytorch and trained on 2 NVIDIA V100 GPUs. For the base network trained on ImageNet we use SGD with momentum as the optimizer. We set the momentum rate to be 0.9, the initial learning rate to be 0.1 and use a batch size of 256. We train the network with a total of 120 epochs and the learning rate decays twice at 80th and 100th epoch with a factor of 10. To prevent overfitting, we use a weight decay (L2 regularization) rate of 0.0001.

For the multi-domain extension of depthwise separable convolution, we keep the same optimization settings as training the base network. We train the network with a total of 100 epochs and the learning rate decays twice at 60th and 80th epoch by a factor of 10. We apply weight
decay (L2 regularization) to prevent overfitting. Since the size of the datasets are highly unbalanced, we use different weight decay parameters for different domains. Similar to \cite{rebuffi18efficient}, higher weight decay parameters are used for smaller datasets. In particular, 0.002 for DTD, 0.0005 for Aircraft, CIFAR100, Daimler pedestrain, Omniglot and UCF101, and 0.0003 for GTSTB, SVHN and VGG-Flowers. 

For soft sharing, we train the network with a total of 10 epochs and the learning rate decays once at the 5th epoch with a factor of 10. Other settings are kept the same as training multi-domain models.

\section{Results and Analysis}

\subsubsection{Quantitative Results} ~\\
The results of the proposed approach and the baselines on Visual Decathlon Challenge are shown in Table \ref{table:results}. Our approach achieves the highest score among all the methods while requiring the least amount of parameters. In particular, the proposed approach improves the current state-of-the-art approaches by 100 points with only 50\% of the parameters. The ResNet-26 with depthwise separable convolution surpasses the performance of the original ResNet-26 by a large margin on ImageNet (63.99 vs 60.32). On other smaller datasets, our approach still achieves better or comparable performance to the baselines. The improvement can be attributed to  the sharing of pointwise convolution that has a regularization effect and allows the training signals in ImageNet to be reused when training new domains.

Compared with other variations of the modified ResNet-26, our approach still achieves the highest score. Our approach obtains a remarkable improvement (3507 vs 2756) with only 20\% of the parameters compared with Individual Network. One reason for the improvement is that the proposed approach is more robust to overfitting, especially for some small datasets. While only training domain-specific classifier layers leads to the smallest model, the score is about 1000 points lower than the proposed approach. Compared with Depthwise Sharing, the assumption of sharing pointwise convolution leads to a more compact and efficient model (3507 vs 3234). This validates our assumption that it is preferable to share pointwise convolution rather than depthwise convolution in the setting of mutli-domain learning. We provide more qualitative results in the next section to support this claim. 

 \begin{figure*}[ht]
\centering
\includegraphics[align=c,width=.04\textwidth]{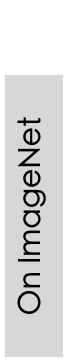} 
\includegraphics[align=c,width=.029\textwidth]{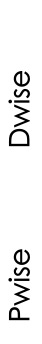} 
\includegraphics[align=c,width=.9\textwidth]{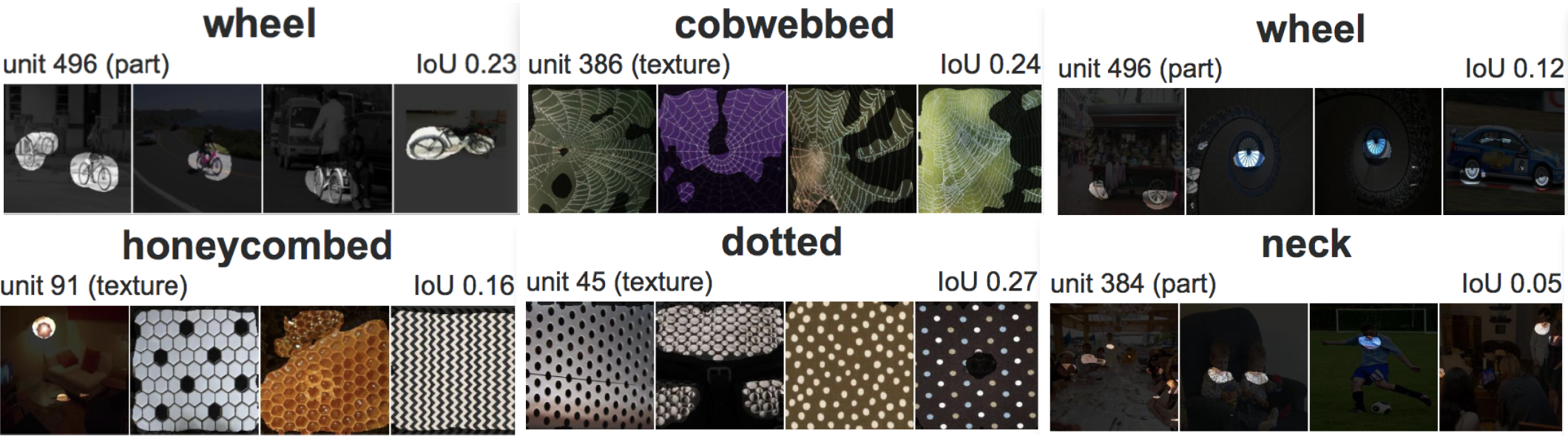} 

\includegraphics[align=c,width=0.8\textwidth]{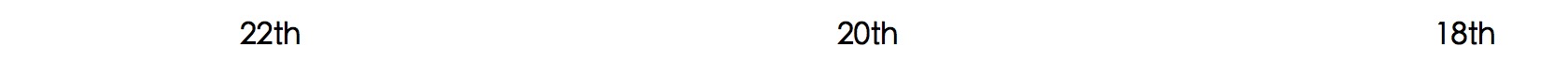}

\includegraphics[align=c,width=.04\textwidth]{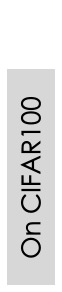} 
\includegraphics[align=c,width=.029\textwidth]{label.jpg} 
\includegraphics[align=c,width=.6\textwidth]{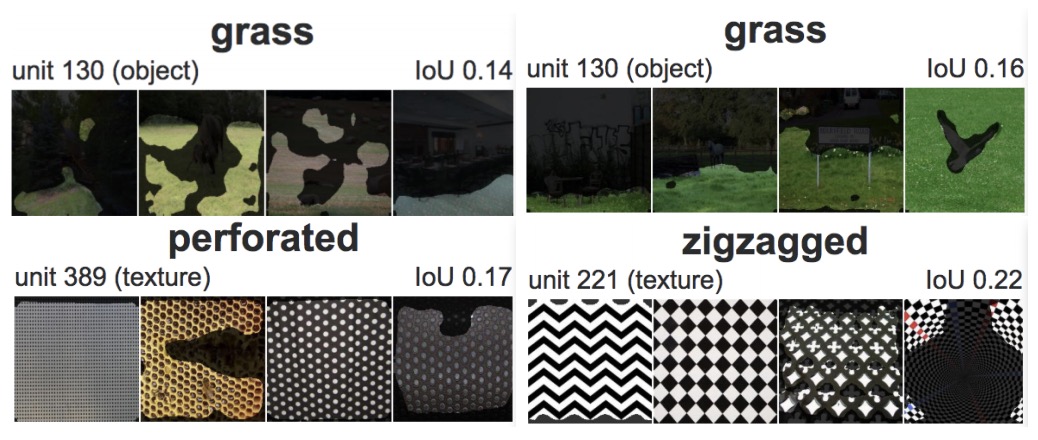} 

\includegraphics[align=c,width=0.75\textwidth]{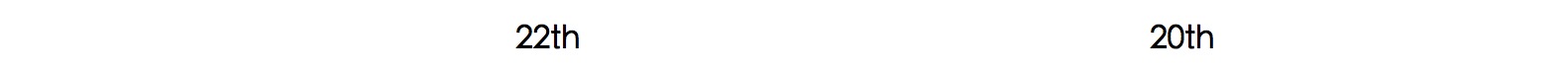} 

\caption{A comparison of visual concepts identified by network dissection in ResNet-26 with depthwise separable convolution trained on ImageNet and CIFAR100. The first two rows demonstrate the results on ImageNet and the last two rows demonstrate the results on CIFAR100. The columns show the results in different layers. The highest-IoU matches among hidden units of each layer are shown. The hidden units of the pointwise convolution in the 18th layer detect no visual concepts. }
\label{fig:concepts}
\end{figure*}

\begin{figure}[h]
\centering
   \begin{subfigure}{0.49\linewidth} \centering
     \includegraphics[scale=0.32]{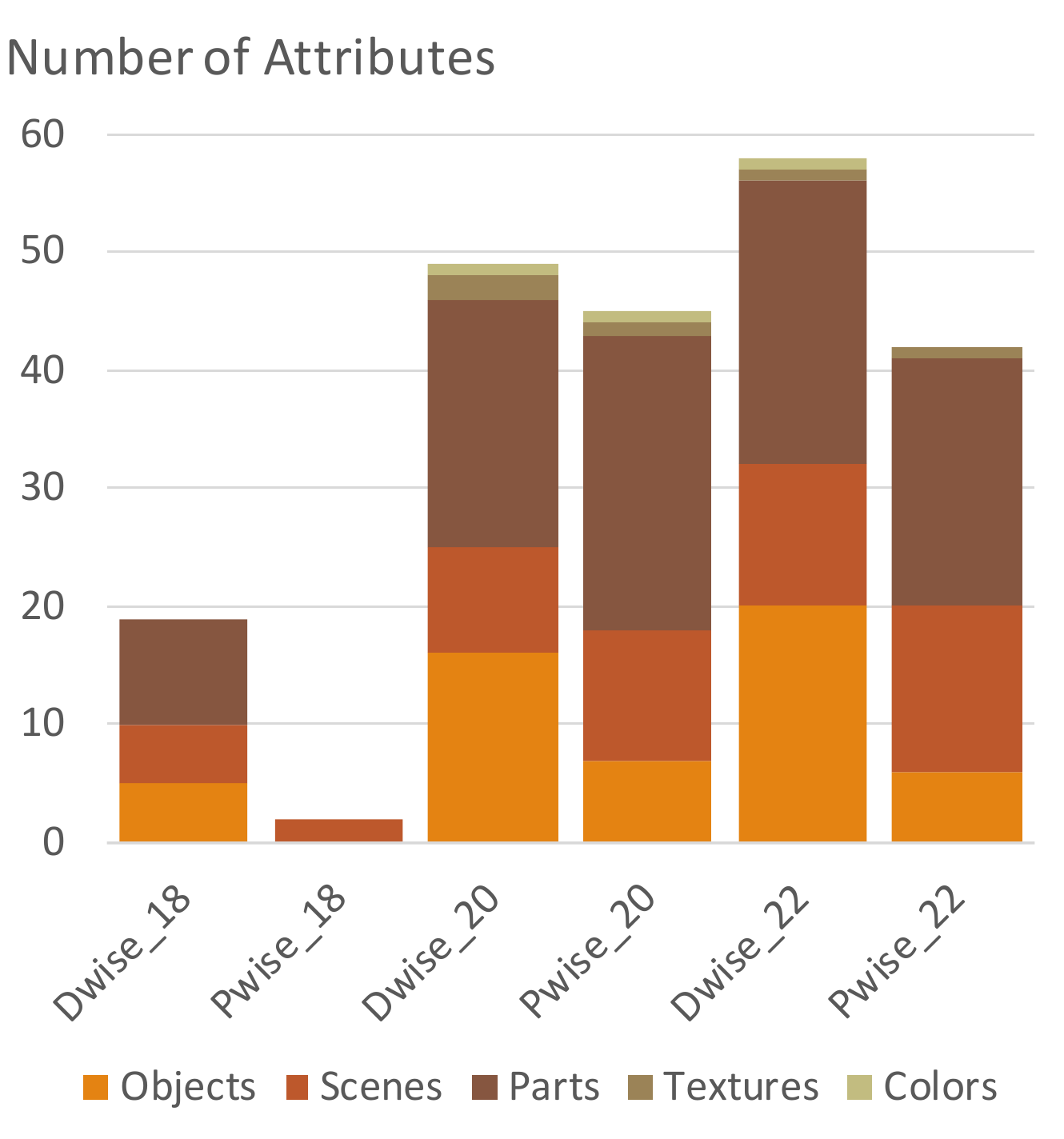}
     \caption{On ImageNet}
   \end{subfigure}
   \begin{subfigure}{0.49\linewidth} \centering
     \includegraphics[scale=0.32]{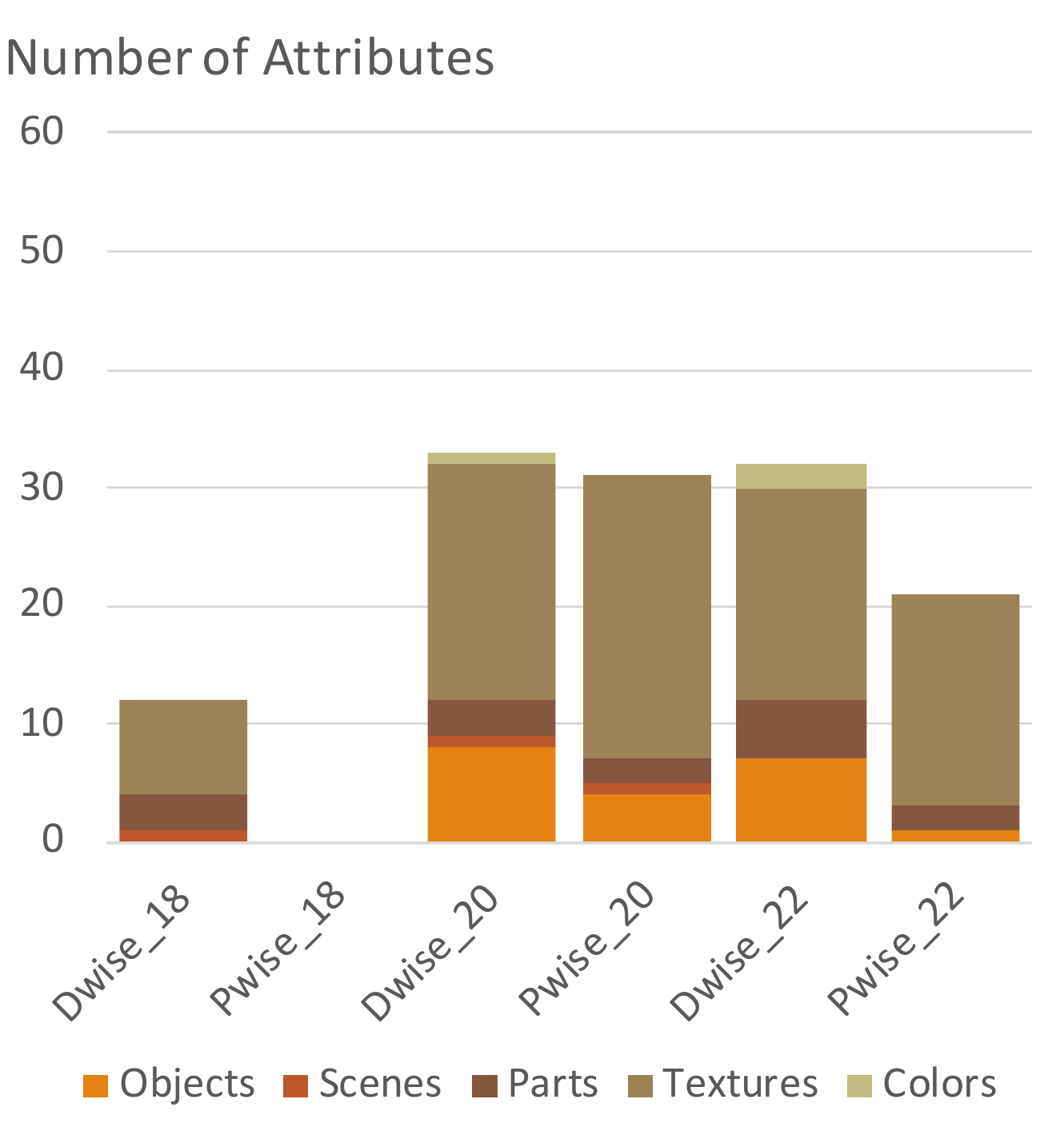}
     \caption{On CIFAR100}
   \end{subfigure}
\caption{Number of attributes captured by the hidden units of depthwise convolution and pointwise convolution in the 18th, 20th and 22th convolutional layer.}   \label{fig: concepts_num}
\end{figure}

\subsubsection{Qualitative Results} ~\\
This section presents our visualization results of deptwise convolution and pointwise convolution based on \textit{network dissection} \cite{bau2017network}. \textit{Network dissection} is a general framework for quantifying the interpretability of deep neural networks by evaluating the alignment between individual hidden units and a set of semantic concepts. The accuracy of unit $k$
in detecting concept $c$ is denoted as $IoU_{k,c}$. If the value of $IoU_{k,c}$ exceeds a threshold then we consider the unit $k$ as a detector for the concept $c$. The details of calculating $IoU_{k,c}$ is omited due to space limitation. 

In the experiments, we use the individual networks trained on ImageNet and CIFAR100 as examples. We visualize the hidden units in the 18th, 20th, 22th convolutional layers. Fig \ref{fig:concepts} shows the interpretability of units of the depthwise convolution and pointwise convolution in the corresponding layer. The highest-IoU matches among hidden units of each layer are shown. We observe that the hidden units in depthwise convolution detect higher level concepts than the units in pointwise convolution. The units in the depthwise convolution can capture part or object while the units in pointwise convolution can only detect textures. Moreover, Fig \ref{fig: concepts_num} shows the number of attributes captured by the units in depth convolution and pointwise convolution. The results demonstrate that depthwise convolution consistently detects more attributes than pointwise convolution. These observations imply that pointwise convolution are mostly used for capturing low level features which can be generally shared across different domains.

\subsubsection{Soft Sharing of Trained Depthwise Filters}
Table \ref{Table: transfer} shows the results of soft sharing. Regardless of the different placements of the softmax gate, the base approach without sharing still achieves the highest score on Visual Decathlon Challenge. One possible reason is that the datasets are from very different domains, sharing information between them may not generally improve the performance. However, for some specific datasets, we still observe some improvement. In particular, by sharing early layers we can obtain a slightly higher accuracy on DTD and SVHN. Another observation is that sharing later layers leads to a higher score than other alternatives. This implies that although images in different domain may not share similar low level features, they can still be benefited from each other by transfering information in later layers.

\section{Conclusion}
In this paper, we present a multi-domain learning approach based on depthwise separable convolution. The proposed approach is based on the assumption that images from different domains share the same channel-wise correlation but have domain-specific spatial-wise correlation. We evaluate our approach on Visual Decathlon Challenge and achieve the highest score among the current approaches. We further visualize the concepts detected by the hidden units in depthwise convolution and pointwise convolution. The results reveal that depthwise convolution captures more attributes and higher level concepts than pointwise convolution.

\section{Acknowledgment}
Work done during internship at IBM Research. This work is supported in part by CRISP, one of six centers in JUMP, an SRC program sponsored by DARPA. This work is also supported by NSF CHASE-CI \#1730158.

\bibliographystyle{aaai}

\end{document}